\documentclass{article}
\usepackage{spconf,amsmath,graphicx}
\usepackage{spconf}
 \usepackage{algorithm}
\usepackage{algpseudocode}
\usepackage{footnote}
\usepackage{placeins}
\usepackage{amssymb}
\usepackage{amsmath}
\usepackage{textcomp}
\usepackage{multirow}
\usepackage{placeins}
\usepackage{footnote}
\usepackage{tabulary}
\usepackage[table,xcdraw]{xcolor}
\usepackage{cite}

\title{SalProp: Salient object proposals via aggregated edge cues}
\name{Prerana Mukherjee$^{\dagger}$ \qquad Brejesh Lall$^{\dagger}$ \qquad Sarvaswa Tandon$^{\star}$}

\address{$^{\dagger}$ Indian Institute of Technology Delhi, India,
    $^{\star}$ National Institute of Technology Goa, India}
\providecommand{\keywords}[1]{\textbf{\textit{Index terms---}} #1}
\begin{document}
%
\maketitle
\begin{abstract}
In this paper, we propose a novel object proposal generation scheme by formulating a graph-based salient edge classification framework that utilizes the edge context. In the proposed method, we construct a Bayesian probabilistic edge map to assign a saliency value to the edgelets by exploiting low level edge features. A Conditional Random Field is then learned to effectively combine these features for edge classification with object/non-object label. We propose an objectness score for the generated windows by analyzing the salient edge density inside the bounding box. Extensive experiments on PASCAL VOC 2007 dataset demonstrate that the proposed method gives competitive performance against $10$ popular generic object detection techniques while using fewer number of proposals.
\end{abstract}


\section{Introduction}
\label{sec:intro}
 
Humans have an excellent ability to simultaneously localize, detect and recognize objects. For machines to know the exact spatial extent of the objects, sufficient training from various exemplar models is required and involves meticulous selection of the object parts from potentially confusing background knowledge. Given the image space, the plausible set of object hypotheses is exponentially large. To select the correct subset of \textit{'good'} object regions and provide a tight bound on the spatial limit of the bounding box involves appropriate feature selection. Thus, the key solution to effective object proposal generation is to leverage the strength of feature statistics. Although with the advent of deep learning based techniques \cite{sermanet2014overfeat, girshick2015fast} and the availability of huge corpus of image data the task of training a machine with huge manually annotated data has eased a lot. Still, it is difficult to capture many interesting patterns like convexity and smoothness of region boundaries locally. There is a scope of improvement for appearance of a new object category. Therefore, the need arises for a model which captures the essence of likeliness of the object regions to provide a suitable set of object proposals\cite{alexe2012measuring, uijlings2013selective, rahtu2011learning, manen2013prime, rantalankila2014generating, arbelaez2014multiscale, qi2015making}. 

Another approach to object localization is the generic object region proposal strategy \cite{pont2017multiscale, rantalankila2014generating, humayun2014rigor}.  Segmentation based on regions is more appealing in the sense that the regions inherently contain the shape and scale information about the objects. There is minimal hindrance in terms of background clutter. But, it is extremely difficult to generate coherent non-overlapping segments. So, rather an efficient scheme for generating few window based proposals having a tight coverage on the object is more convenient and logical for applications like classification \cite{uijlings2013selective, zhou2016dave}, video summarization , segmentation \cite{uijlings2013selective}, action recognition.\\
Recently, a couple of techniques have tried to exploit the potential of edges as an object localization cue \cite{zitnick2014edge,qi2015making}. Edges capture most of the shape information thus preserving important structural properties contained in the image. They often occur at locations adhering to the object boundaries which make them a suitable candidate as precursor to object localization as well as segmentation. Major advantage of proposed technique in contrast to to \cite{zitnick2014edge,qi2015making} is that an inherent saliency ordering is preserved in the set of generated proposals apart from providing high precision and recall rates even with lesser number of proposals.  Generating fewer number of high precision proposals also reduces the number of spurious false positives in the detection \cite{alexe2012measuring}. The contemporary deep learning based methods provide excellent results but require huge amount of training data and sometimes initialization with \textit{good} object hypotheses \cite{girshick2015fast}. Our technique can be augmented with such techniques as well. In particular, we demonstrate in the experiments that even with $\sim10-20$ object proposals the detection rate is quite high ($53\%-61.55\%$ at IoU=$0.5$). In view of the above discussion, the key contributions can be summarized as,
\begin{enumerate}
\item To the best of our knowledge, this is the first work to establish the concept of object edge classification in a Conditional Random Field (CRF) framework for object proposal generation. 
\item We demonstrate good performance (high recall rates) utilizing very few number of object proposals. We rank the key objects in relative order of salience based on the edge saliency by the proposed scheme. 

\end{enumerate}

\section{Proposed Methodology}
\label{sec:methodology}

\begin{figure}[hbtp]
\centering
{
\includegraphics[scale=0.38]{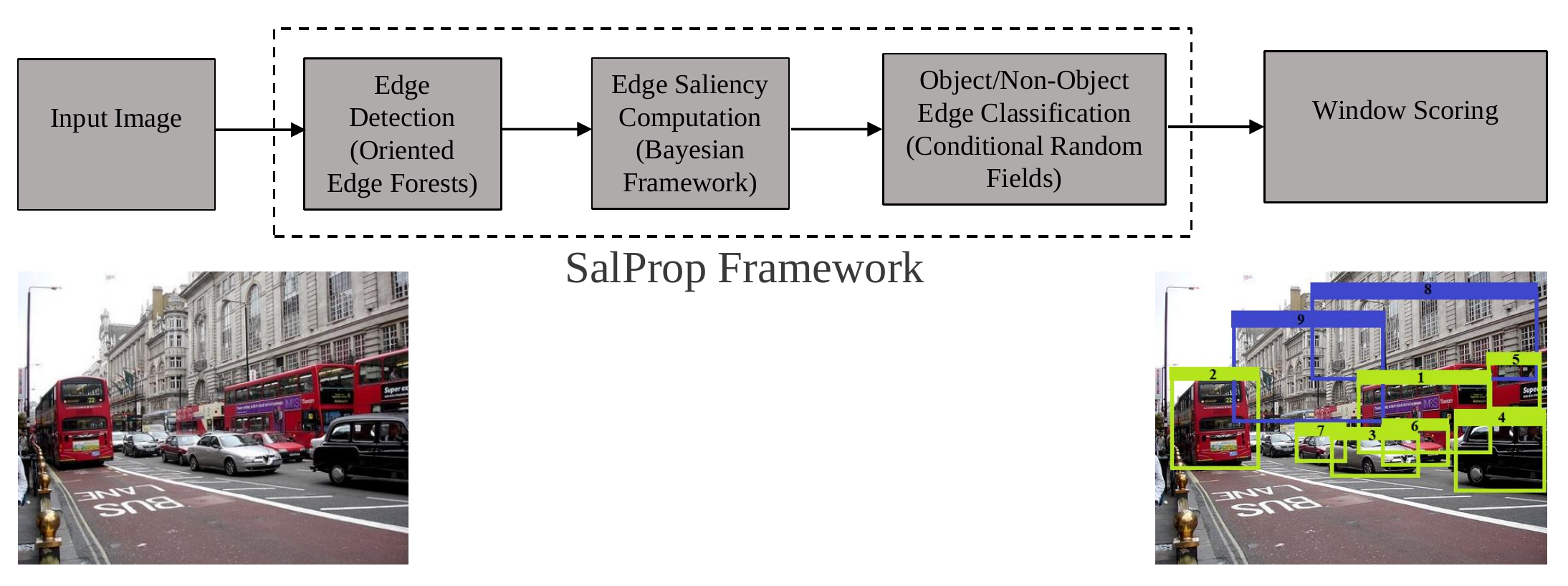}
\caption{The SalProp Framework. Given any RGB image, we
generate proposals ranked in the order of saliency. Green boxes contain the most salient objects having higher rank and blue boxes
contain less salient objects and are ranked lower in the proposal set. The number assigned to each box indicates its saliency ranking in the proposal pool.}
\label{fig:workflow}
}
\end{figure}

In this section, we give a detailed overview of the proposed salient object proposal generation scheme. The end-to-end pipeline of the proposed method is shown in Fig. \ref{fig:workflow}.

\subsection{Edge Saliency Computation (Bayesian Framework)}
\label{sec:edgesaliency}

\begin{figure}[hbtp]
\centering
{
\includegraphics[scale=0.43]{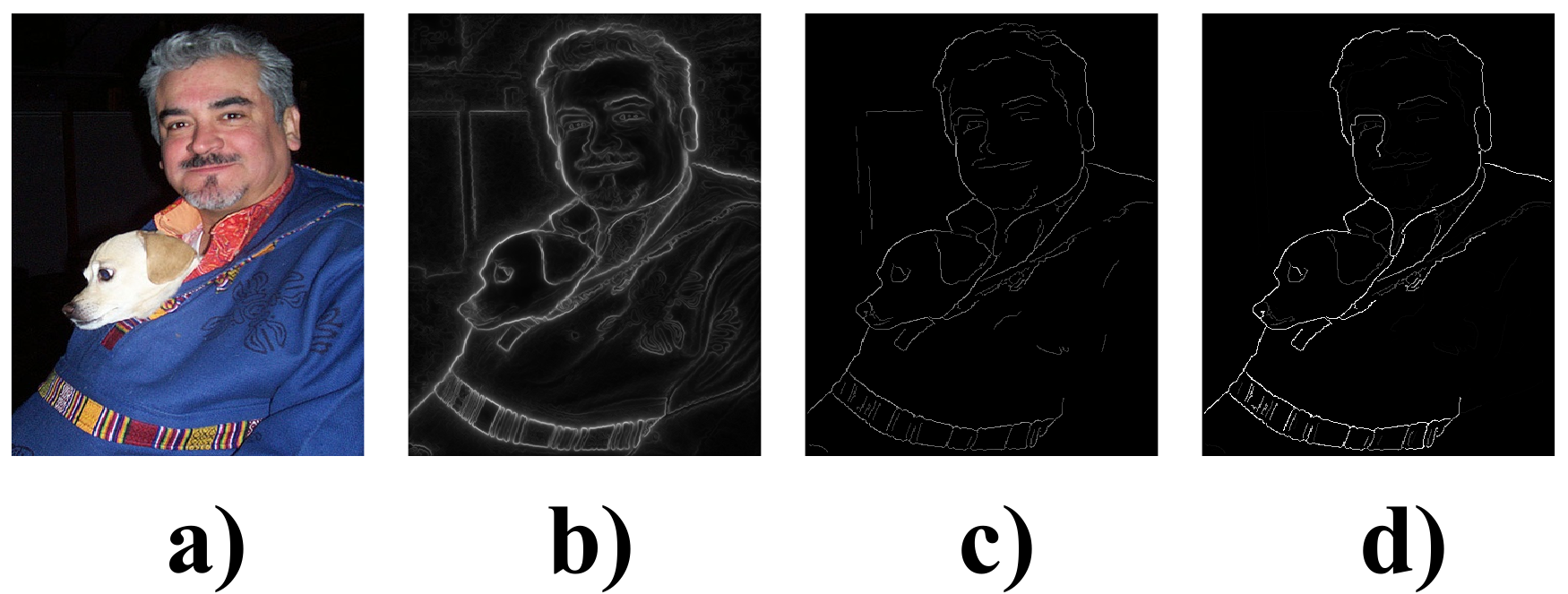}
\caption{(a) Original image (b) Edge map using OEF (c) After NMS and thresholding (d) Bayesian Probabilistic edge map (indicating saliency of edge segments)}
\label{fig:probsal}
}
\end{figure}

The early processing units in the primate visual system help in detecting the object edge responses which are then perceptually grouped to form continuous contours. Deriving motivation from this, we describe the strategy for identification of edge pixels corresponding to the objects so that this edge map can be used as a strong prior for object localization. To this end, we utilize a sparse edge map to form a probabilistic saliency map in which each edgelet (edge segment) is assigned a saliency value, thus providing it a distinctiveness score. The score is computed by encoding the local edge context information i.e. texture, color gradient, edge magnitude. We pose the edge saliency detection as a Bayesian inference problem to indicate the edge segments belonging to the object (salient) or background (non-salient). We estimate the prior distribution of salient or background edges based on their edge magnitude since stronger edges are more likely to be a part of an object. Given an image, we first compute the edge responses with the Oriented Edge Forests (OEF) boundary detector \cite{hallman2015oriented} which is highly efficient in detecting object boundaries and computationally less expensive. We utilize the sparse variant of OEF detection in which non-maximal suppression (NMS) is used. The resultant sparse edge map consists of each pixel $i$ having an edge magnitude $|e_i|$. We further perform a thresholding (provides computational efficiency) by considering edge segments with length $l>15$ and edge pixels having magnitude $|e_i|>$40. These values provided best results in our experimental analysis. The posterior probability of each edge segment denoted by $p(sal|\mathfrak{s})$ having a relative edge strength $\mathfrak{s}$ in the sparse edge map is mathematically formalized as:
\begin{equation}
\small{p(sal|\mathfrak{s})=\frac{p(sal)p(\mathfrak{s}|sal)}{p(sal)p(\mathfrak{s}|sal)+p(bg)p(\mathfrak{s}|bg)},}
\end{equation}
where $p(sal|\mathfrak{s})$ is the probability of the edge segment being salient. $p(sal)$ and $p(bg)$ are the prior probabilities of the edge segment to be salient (object edges) or background respectively. $p(\mathfrak{s}|sal)$ and $p(\mathfrak{s}|bg)$ are the likelihood of observations. $\mathfrak{s}$ denotes the relative edge strength as computed in Eq. \ref{eq:edgestrength}. Edge saliency prior of $j^{th}$ edge segment is computed as:
\begin{equation}
\small{p(sal) = \frac{\mathfrak{N}}{max_j \mathfrak{N}_j}, \mathfrak{N}=f_{G}.f_{LTP}.\mathfrak{s},}
\end{equation}
where $\mathfrak{N}$ indicates the scalar multiplication of the texture, color and edge magnitude values of the edge pixels in the $j^{th}$ edge segment.
We integrate the magnitudes of color
gradients of a particular orientation ($G_{o,i}$), $o\epsilon\{0^\circ, 45^\circ, 90^\circ, 135^\circ\}$ along the edges denoted by $f_{G}$, given as:
\begin{equation}
\small{f_{G}=\sqrt{\sum_o(\sum_i G_{o,i})^2}\\.}
\end{equation}
$f_{LTP}$ is the local ternary pattern (LTP) of the edge pixels $I_i$ contained in the $j^{th}$ edge segment is computed by comparing the intensity value of it with the intensity values of its neighbors denoted by $I_{nb}$ using a kernel of size $3$. In \cite{tan2010enhanced}, the authors utilize the LTP code as a combination of its “upper” and “lower” local binary pattern (LBP) codes. Since, we represent LTP for the edge segments only we take the average variance of this combination over the edge. Here, $T$ is user defined threshold and $B=8$. We take the variance of all the LTP values of the edge pixels for a particular segment given as:
\begin{align}
& \small{f_{LTP}=\frac{\sigma(ULBP)+\sigma(LLBP)}{2},}\\
& \small{ULBP=\sum_{b=0}^{B-1}s'(I_{nb}-I_{i}).2^b,}\\ 
& \small{LLBP=\sum_{b=0}^{B-1}f'(I_{nb}-I_{i}).2^b,}\\
& \small{s'(z)=\left\{\begin{matrix}
1 & z \geq T \\ 
0 & otherwise 
\end{matrix}\right.f'(z)=\left\{\begin{matrix}
1 & z \leq -T \\ 
0 & otherwise 
\end{matrix}\right.}
\end{align}
The maximum magnitude value $\mathfrak{s}$ of edge pixels in $j^{th}$ edge segment is computed as follows:
\begin{equation}
\label{eq:edgestrength}
\small{\mathfrak{s}=max_i(|e_i|).}
\end{equation}
The background prior is given as,
\begin{equation}
\small{p(bg)=1-p(sal).}
\end{equation}
To find the likelihood, we need to separate the edge segments into salient or background segments. If the edge magnitude $\geq \beta.\mathfrak{s}$, we consider it as salient, else it is a background edge segment. Here, $\beta$ indicates the edge magnitude threshold, where $\beta >0$. We then compute the normalized histograms $h_s$ and $h_{bg}$ of the edge magnitudes of the edge pixels in salient and background edge segments respectively with $10$ bins each. The observation likelihoods $p(\mathfrak{s}|sal)$ and $p(\mathfrak{s}|bg)$ are calculated from $h_s$ and $h_{bg}$ respectively based on bin value to which $\mathfrak{s}$ of the edge segment belongs. The probabilistic edge map is shown in Fig. \ref{fig:probsal}.

\subsection{CRF for Edge Classification}
\label{sec:crf}

We formulate Edge Feature Graph Conditional Random Field \cite{lafferty2001conditional} (CRF) model to learn the conditional distribution over the edge segment labeling given an image using the local edge context. CRF is used here for structured prediction for the edge labeling problem. The links in the graph are made between the edge segments (nodes) which are spatially close. The nodes are associated with $7$-D feature vector (Section \ref{subsubsec:salientedgefeat}). The score associated with each link is denoted as $e_{ij}$ given by a 4-D feature vector [Up/Down, Right/Left, mean, variance]. The first two elements ($0/1$) in the vector denote the relative position of node $i$ with respect to node $j$. The next two elements denote the mean and variance in the feature differences between the two nodes in the graph.
The objective function (energy) of the structured prediction is given as:
\begin{equation}
\small{E(L|X)=\sum_{i\epsilon \mathcal{V}} \boldsymbol{\phi}(l_i,X; \mathbf{W_1})+\sum_{\{i,j\}\epsilon\mathcal{E}}\boldsymbol{\psi}(l_i,l_j,X;\mathbf{W_2}),}
\end{equation}
where $L$ is the structured label, $X$ is the structured input features, $l_i$ is the label of the node, $\boldsymbol{\phi}(l_i,X; \mathbf{W_1})$ are unary potentials and $\boldsymbol{\psi}(l_i,l_j,X;\mathbf{W_2})$ indicates pairwise potentials. The objective function is optimized using Block-coordinate Frank Wolfe Structured SVM to compute $\bf{W}=[\mathbf{W_1} \quad \mathbf{W_2}]$.

\subsubsection{Local Edge Features}
\label{subsubsec:salientedgefeat}

We consider two image patches (radius=$5$) on either side of the edge segment to take into account the contextual information around the edge. We uniformly sample half of the data points (pixels) in region $A_1$ and $A_2$ to avoid overfitting. We next compute the texture features for the data points in these regions. For this, we compute a $5$-dimensional filter bank at scale $k$\footnote{We use perceptually uniform CIELab color space. The filter bank consists of Difference of Gaussian (DoG) at $2$ scales \{$k$, 2$k$\}, Laplacian of Gaussian (LoG) at $3$ scales \{$k$, 2$k$, 4$k$\}. These filters are applied only to the luminance channel. (${k}$ taken as $0.5$)}. We next compute the variance of the DoG and LoG feature vectors of each region. We concatenate the feature vectors in ascending order of variance as [$DoG_1$, $DoG_2$] and [$LoG_1$, $LoG_2$]. The intuition behind this is that the region having low texture variation is likely to belong to object region and vice-versa while maintaining an ordering for CRF training. Thus, the $7$-D feature vector for each edge segment is represented by the vector, [$f_G$, $DoG_1$, $DoG_2$, $LoG_1$, $LoG_2$, $f_{LTP}$, $\mathfrak{s}$]. The computation of $f_G$, $f_{LTP}$ and $\mathfrak{s}$ has been explained in Section \ref{sec:edgesaliency}.

\subsection{Window Generation and Scoring}
\label{sec:winscore}

We proceed with a sliding window technique for proposal generation over position, scale and aspect-ratio. Each successive window maintains an Intersection over Union (IoU) with the previous window and the step size is calculated accordingly. The IoU is taken as $0.65$ (as in EdgeBoxes approach \cite{zitnick2014edge}). Scale is set from $0.5\%$ to $95\%$ of the image size with $1\%$ increment between scales. The aspect ratio ranges from $1/3$ to $3$. All edge segments that fall completely inside the proposal window increase the score depending on their edge length and saliency value. Furthermore, the score discourages larger windows to have high scores by dividing the score by the area of the window given as,
\begin{equation}
\small{S_w=\frac{\sum_j s_j.l_j}{\sqrt[]{Area_w}},}
\end{equation}
where $s_j$ indicates the saliency value and $l_j$ is the length of the $j^{th}$ edge segment. $Area_w$ is the area of window $w$. There are two
necessary post processing steps for generating better proposals: Refinement and Non-Maximal Suppression (NMS). We perform these steps in congruent lines to those in \cite{zitnick2014edge}.
\begin{table*}[]
\footnotesize
\centering
\caption{Comparison of top 1000
proposals with state-of-the-art techniques on AUC\% (higher the better), number of proposals (N) at 75\% recall (lower the better) and recall\% (higher the better). '-' indicates that the particular recall rate is not reached.}
\label{table1}
\begin{tabular}{|l|l|l|l|l|l|l|l|l|l|l|}
\hline
\multirow{2}{*}{\textbf{Method}} & \multicolumn{3}{c|}{\textbf{IoU=0.5}}             & \multicolumn{3}{c|}{\textbf{IoU=0.6}}             & \multicolumn{3}{c|}{\textbf{IoU=0.7}}             & \multirow{2}{*}{\textbf{Time(in s)}} \\ \cline{2-10}
                                 & \textbf{AUC} & \textbf{N@75\%} & \textbf{Recall} & \textbf{AUC} & \textbf{N@75\%} & \textbf{Recall} & \textbf{AUC} & \textbf{N@75\%} & \textbf{Recall} &                                         \\ \hline
EdgeBoxes70 \cite{zitnick2014edge}            &   65.82            & 86                &   93.45              &   60.52           &  141               &  90.73               &   53.03           & 294                 &   84.15             &   0.25 
                                     \\ \hline
PE \cite{qi2015making}         &1.8             &  -               & 10.4               &  0.08           & -                & 4.7                &  0.02            &  -               & 1.2               & 7.2                                        \\ \hline
MCG \cite{pont2017multiscale}                    & 71            & 37                &  94.6               & 62.8             &  95               & 90.2                & 62.5             &  366               & 83               &  34                                       \\ \hline
Objectness \cite{alexe2012measuring}             &62             &  145               &    89             &   52           &   504              &  78               &  30           &  -               &  41             &   3                                      \\ \hline
Rahtu \cite{rahtu2011learning}                   & 57             &   278              &    84            &   50           & 551                & 79               &  43.5            &  -               & 73.5                &  3
                                       \\ \hline
RP \cite{manen2013prime}                    &59.3              &  129               &  89               & 50             & 315                & 83                &   40.7           &  1000               &  75              &   1                                      \\ \hline
Rantalankila \cite{rantalankila2014generating}            &  25.14            &  511               &  86.38              & 21.63              & 718               & 79.77                &  17.76            &   -              &70.75                &  10                                       \\ \hline
SS \cite{uijlings2013selective}                      &62.3              & 105                &93                 &  54            &   207              &  88              &   45.3           &  544               &  80              &   10
                                      \\ \hline
Rigor\cite{humayun2014rigor}                   &40.39             &  -               &  67.43               &    32.05          &   -              &  54.5               &   23.44          &   -              & 40.73                &   6.84                                      \\ \hline
GOP\cite{krahenbuhl2014geodesic}                     &47.8               &   155              &    93             &  41            &   272              &  87              &  33.4           &    705             &  76              & 0.9
                                        \\
\hline
\textbf{SalProp}                     &67.5              &      74           &  91               &  58.1           &   244              &  84               &   44           &  -               &  71.3               &  7                                       \\
\hline
\end{tabular}
\end{table*}

\section{Experimental Results}
\label{sec:results}

We utilize Pystruct 0.2.5 structured prediction \cite{muller2014pystruct} for implementing CRF model. The CRF model is trained on the MSRA1000 saliency dataset\cite{achanta2009frequency} which has been chosen due to higher distinction of edge features between the object and background. The training is performed in two steps. First, the edgemap is extracted using OEF followed by NMS and thresholding. Next, we perform $k$-means clustering on edge magnitude of edges (with $k$=2) to segregate them into object and non-object edges. We take the ground truth edges and higher magnitude edges as object edges while lower magnitude edges as non-object edges. CRF is trained by utilizing the edge features as discussed in Section \ref{subsubsec:salientedgefeat}. The model is further evaluated on PASCAL 2007 \cite{everingham2010pascal} with 2510 validation set images (to get the final parameter setting) and 4952 testing images. The parameter setting used in Section \ref{sec:edgesaliency} involves $T$ taken as $5$ and $\beta=0.8$.

\subsection{Quantitative Evaluation}

\begin{figure}[hbtp]
\centering
{
\includegraphics[scale=0.35]{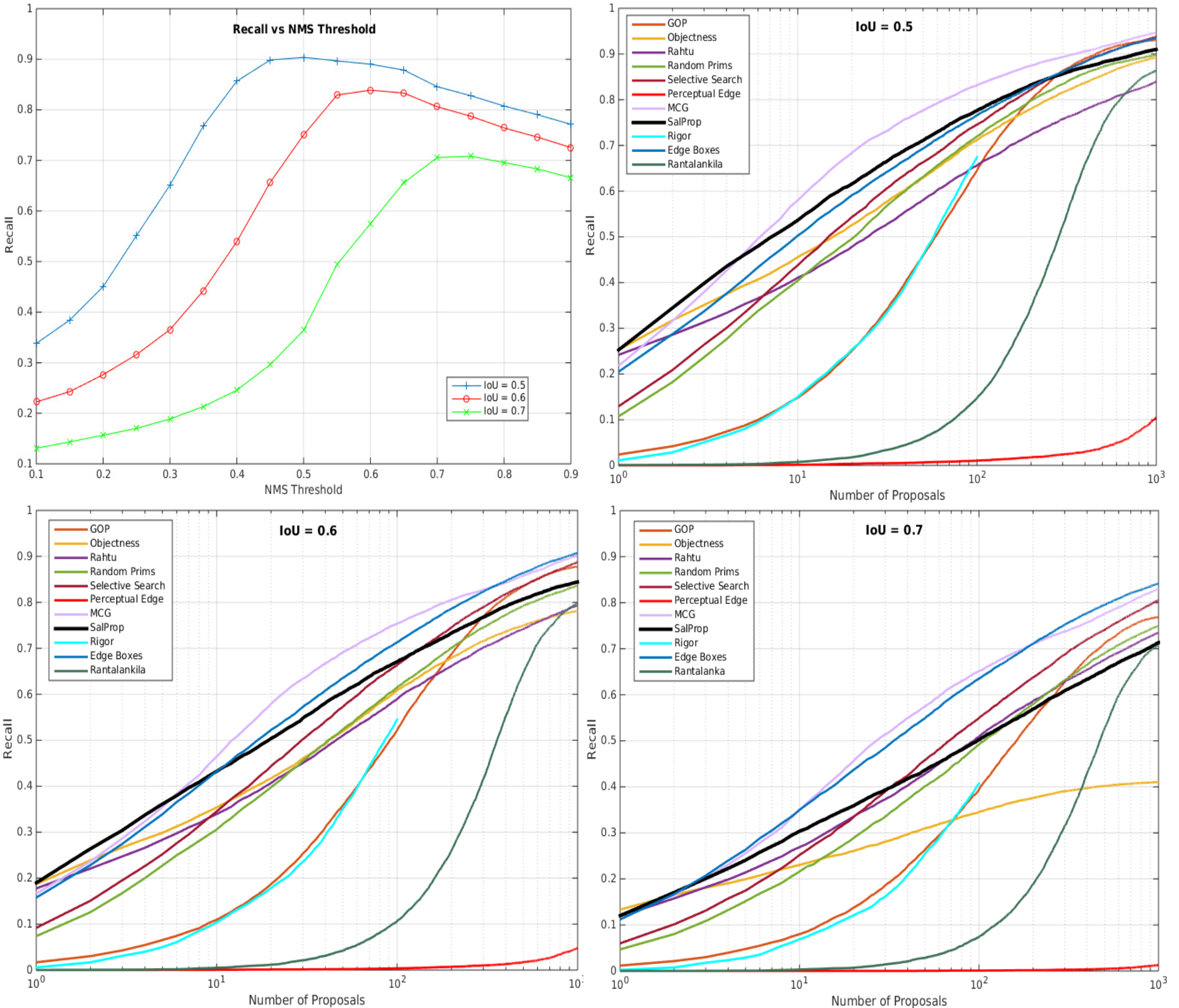}
\caption{(a) NMS cut-off threshold for highest recall value at varying IoU on validation set images. (b)-(d) The detection rate vs. the number of bounding box
proposals for varying IoU = $0.5$, $0.6$ and $0.7$ on validation set images. }
\label{fig:compare}
}
\end{figure}

Table \ref{table1} compares SalProp against the state-of-art algorithms. Fig. \ref{fig:compare}(a) shows cut-off NMS threshold. Fig. \ref{fig:compare}(b)-(d) shows the detection rates when we are varying the number of object proposals at different IoUs. SalProp is the best technique at lower number of proposals achieving over $25\%$ and $19\%$ recall with only $1$ window at IoU=$0.5$ and $0.6$ respectively. At IoU=$0.7$, SalProp outperforms Rahtu\cite{rahtu2011learning} by $3.46\%$, Selective Search\cite{uijlings2013selective} by $5.16\%$, Objectness\cite{alexe2012measuring} by $7.32\%$, Randomized Prim's\cite{manen2013prime} by $8.71\%$, GOP\cite{krahenbuhl2014geodesic} by $22.36\%$, Rigor\cite{humayun2014rigor} by $23.46\%$, Rantalankila\cite{rantalankila2014generating} by $30.05\%$ and Perceptual Edge\cite{qi2015making} by $30.35\%$ at top-$10$ proposals demonstrating that it consistently ranks higher the object proposals that are closer to the ground truth when lower number of proposals are considered.

MCG\cite{pont2017multiscale} and EdgeBoxes\cite{zitnick2014edge} techniques outperform SalProp at a few number of proposals. SalProp provides comparable performance to EdgeBoxes while having a computational speedup of $5$x over MCG (Table \ref{table1}) which is based on learning based setting whereas SalProp operates in a computationally efficient no explicit learning based setting. The results demonstrate that the proposed algorithm performs better on varying IoU thresholds for less number of candidate proposals while maintaining high recall at higher proposals. The important note to make here is that except Objectness  the compared approaches do not take into account the saliency aspect of an object which is a key property in characterizing an object \cite{mukherjee2017saliency}. Our method outperforms objectness by $2\%$, $6\%$ and $30\%$ at IoU thresholds $0.5$, $0.6$ and $0.7$ respectively. 
\subsection{Qualitative Evaluation}

\begin{figure}[hbtp]
\centering
{
\includegraphics[scale=.2]{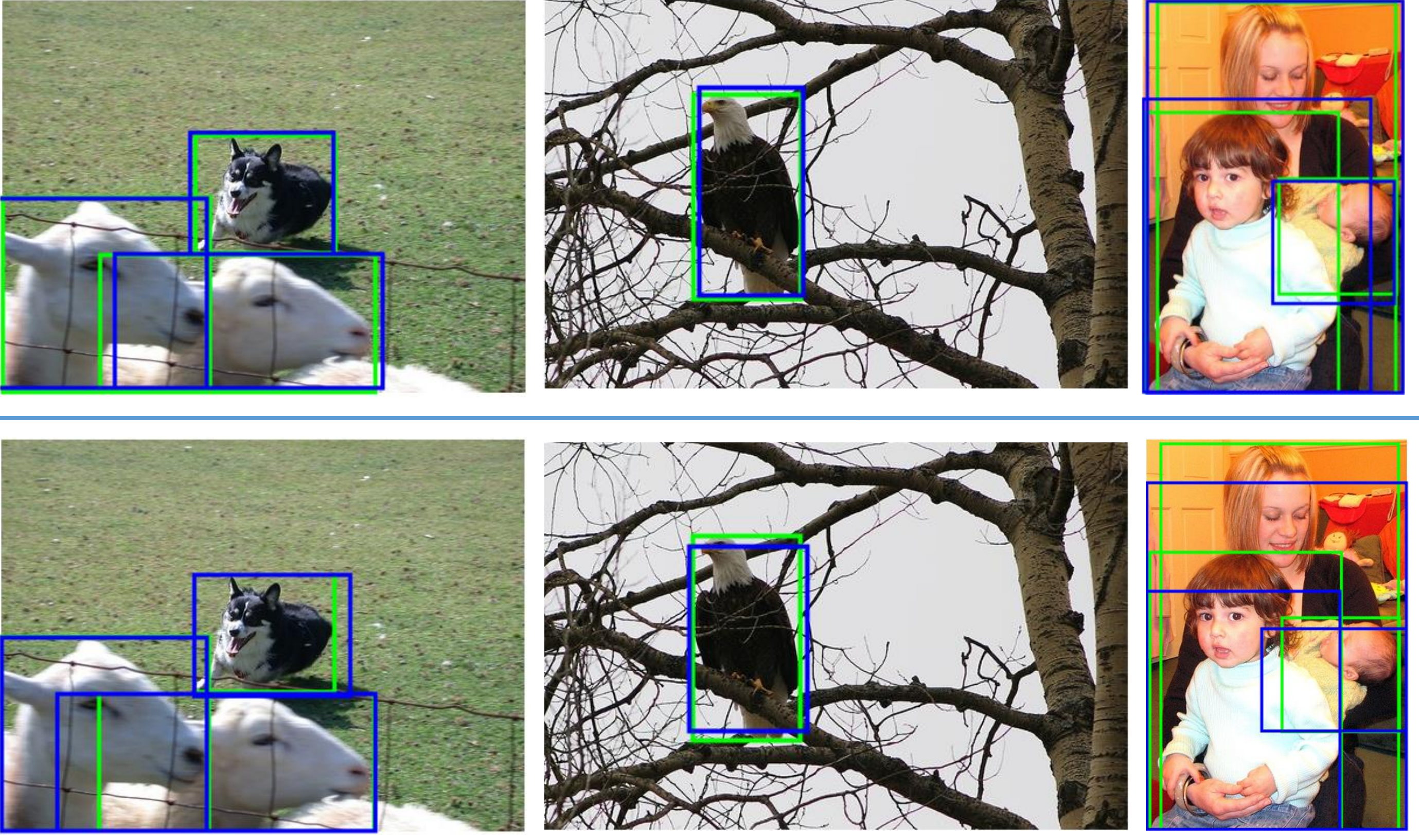}
\caption{Top Row: SalProp, Bottom Row: EdgeBoxes \cite{zitnick2014edge}. Closest bounding boxes (blue) having maximum overlap with the ground truth boxes (green).}
\label{fig:salprop}}
\end{figure}

Fig \ref{fig:salprop} shows qualitative results. The results are computed for IoU=0.7.  
It can be observed that SalProp produces tight bounding boxes (e.g. sheep and babies) and is able to detect occluding and difficult objects with high accuracy. 

\section{Conclusion}
\label{sec:conclusion}

We proposed a novel object proposal generation algorithm which operates in a computationally efficient learning based setting where the salient object edge density inside the bounding box is analyzed to score the proposal set. We provided comprehensive empirical evaluation and comparison with several baselines and existing methods to demonstrate the effectiveness of the technique. We showed that the proposed architecture achieves high recall rates with lesser number of proposals with varying IoU thresholds and subsequently making it more reliable in context of
competing methods. We also ranked the key objects according to their saliency.


\bibliographystyle{IEEEbib}
\bibliography{strings}

\begin{thebibliography}{10}

\bibitem{sermanet2014overfeat}
Pierre Sermanet, David Eigen, Xiang Zhang, Micha{\"e}l Mathieu, Rob Fergus, and
  Yann LeCun,
\newblock ``Overfeat: Integrated recognition, localization and detection using
  convolutional networks,''
\newblock {\em ICLR, 2014}, 2014.

\bibitem{girshick2015fast}
Ross Girshick,
\newblock ``Fast r-cnn,''
\newblock in {\em Proceedings of the IEEE International Conference on Computer
  Vision}, 2015, pp. 1440--1448.

\bibitem{alexe2012measuring}
Bogdan Alexe, Thomas Deselaers, and Vittorio Ferrari,
\newblock ``Measuring the objectness of image windows,''
\newblock {\em IEEE Transactions on Pattern Analysis and Machine Intelligence},
  vol. 34, no. 11, pp. 2189--2202, 2012.

\bibitem{uijlings2013selective}
Jasper~RR Uijlings, Koen~EA van~de Sande, Theo Gevers, and Arnold~WM Smeulders,
\newblock ``Selective search for object recognition,''
\newblock {\em International journal of computer vision}, vol. 104, no. 2, pp.
  154--171, 2013.

\bibitem{rahtu2011learning}
Esa Rahtu, Juho Kannala, and Matthew Blaschko,
\newblock ``Learning a category independent object detection cascade,''
\newblock in {\em 2011 International Conference on Computer Vision}. IEEE,
  2011, pp. 1052--1059.

\bibitem{manen2013prime}
Santiago Manen, Matthieu Guillaumin, and Luc Van~Gool,
\newblock ``Prime object proposals with randomized prim's algorithm,''
\newblock in {\em Proceedings of the IEEE International Conference on Computer
  Vision}, 2013, pp. 2536--2543.

\bibitem{rantalankila2014generating}
Pekka Rantalankila, Juho Kannala, and Esa Rahtu,
\newblock ``Generating object segmentation proposals using global and local
  search,''
\newblock in {\em Proceedings of the IEEE conference on computer vision and
  pattern recognition}, 2014, pp. 2417--2424.

\bibitem{arbelaez2014multiscale}
Pablo Arbel{\'a}ez, Jordi Pont-Tuset, Jonathan~T Barron, Ferran Marques, and
  Jitendra Malik,
\newblock ``Multiscale combinatorial grouping,''
\newblock in {\em Proceedings of the IEEE Conference on Computer Vision and
  Pattern Recognition}, 2014, pp. 328--335.

\bibitem{qi2015making}
Yonggang Qi, Yi-Zhe Song, Tao Xiang, Honggang Zhang, Timothy Hospedales, Yi~Li,
  and Jun Guo,
\newblock ``Making better use of edges via perceptual grouping,''
\newblock in {\em Proceedings of the IEEE Conference on Computer Vision and
  Pattern Recognition}, 2015, pp. 1856--1865.

\bibitem{pont2017multiscale}
Jordi Pont-Tuset, Pablo Arbelaez, Jonathan~T Barron, Ferran Marques, and
  Jitendra Malik,
\newblock ``Multiscale combinatorial grouping for image segmentation and object
  proposal generation,''
\newblock {\em IEEE transactions on pattern analysis and machine intelligence},
  vol. 39, no. 1, pp. 128--140, 2017.

\bibitem{humayun2014rigor}
Ahmad Humayun, Fuxin Li, and James~M Rehg,
\newblock ``Rigor: Reusing inference in graph cuts for generating object
  regions,''
\newblock in {\em Proceedings of the IEEE Conference on Computer Vision and
  Pattern Recognition}, 2014, pp. 336--343.

\bibitem{zhou2016dave}
Yi~Zhou, Li~Liu, Ling Shao, and Matt Mellor,
\newblock ``Dave: A unified framework for fast vehicle detection and
  annotation,''
\newblock in {\em European Conference on Computer Vision}. Springer, 2016, pp.
  278--293.

\bibitem{zitnick2014edge}
C~Lawrence Zitnick and Piotr Doll{\'a}r,
\newblock ``Edge boxes: Locating object proposals from edges,''
\newblock in {\em European Conference on Computer Vision}. Springer, 2014, pp.
  391--405.

\bibitem{hallman2015oriented}
Sam Hallman and Charless~C Fowlkes,
\newblock ``Oriented edge forests for boundary detection,''
\newblock in {\em Proceedings of the IEEE Conference on Computer Vision and
  Pattern Recognition}, 2015, pp. 1732--1740.

\bibitem{tan2010enhanced}
Xiaoyang Tan and Bill Triggs,
\newblock ``Enhanced local texture feature sets for face recognition under
  difficult lighting conditions,''
\newblock {\em IEEE transactions on image processing}, vol. 19, no. 6, pp.
  1635--1650, 2010.

\bibitem{lafferty2001conditional}
John Lafferty, Andrew McCallum, and Fernando Pereira,
\newblock ``Conditional random fields: Probabilistic models for segmenting and
  labeling sequence data,''
\newblock in {\em Proceedings of the eighteenth international conference on
  machine learning, ICML}, 2001, vol.~1, pp. 282--289.

\bibitem{krahenbuhl2014geodesic}
Philipp Kr{\"a}henb{\"u}hl and Vladlen Koltun,
\newblock ``Geodesic object proposals,''
\newblock in {\em European Conference on Computer Vision}. Springer, 2014, pp.
  725--739.

\bibitem{muller2014pystruct}
Andreas~C M{\"u}ller and Sven Behnke,
\newblock ``Pystruct: learning structured prediction in python.,''
\newblock {\em Journal of Machine Learning Research}, vol. 15, no. 1, pp.
  2055--2060, 2014.

\bibitem{achanta2009frequency}
Ravi Achanta, Sheila Hemami, Francisco Estrada, and Sabine Susstrunk,
\newblock ``Frequency-tuned salient region detection,''
\newblock in {\em Computer vision and pattern recognition, 2009. cvpr 2009.
  ieee conference on}. IEEE, 2009, pp. 1597--1604.

\bibitem{everingham2010pascal}
Mark Everingham, Luc Van~Gool, Christopher~KI Williams, John Winn, and Andrew
  Zisserman,
\newblock ``The pascal visual object classes (voc) challenge,''
\newblock {\em International journal of computer vision}, vol. 88, no. 2, pp.
  303--338, 2010.

\bibitem{mukherjee2017saliency}
Prerana Mukherjee and Brejesh Lall,
\newblock ``Saliency and kaze features assisted object segmentation,''
\newblock {\em Image and Vision Computing}, vol. 61, pp. 82--97, 2017.

\end{thebibliography}
\end{document}